\documentclass[conference]{IEEEtran}
\IEEEoverridecommandlockouts
\usepackage{cite}
\usepackage{amsmath,amssymb,amsfonts}
\usepackage{algorithmic}
\usepackage{graphicx}
\usepackage{textcomp}
\usepackage{xcolor}
\usepackage{subcaption}

\newcommand{\h}{\mathbf{h}}
\newcommand{\x}{\mathbf{x}}
\newcommand{\y}{\mathbf{y}}

\newcommand{\U}{\mathbf{U}}
\newcommand{\W}{\mathbf{W}}
\newcommand{\V}{\mathbf{V}}

\newcommand{\R}{\mathbb{R}}

\def\BibTeX{{\rm B\kern-.05em{\sc i\kern-.025em b}\kern-.08em
    T\kern-.1667em\lower.7ex\hbox{E}\kern-.125emX}}
\begin{document}

\title{Sparsity in Reservoir Computing Neural Networks
\thanks{\textbf{This paper is currently under review.} 
This work has been partially supported by the European Union’s Horizon 2020 Research and Innovation program, under project TEACHING (Grant agreement ID: 871385).
https://www.teaching-h2020.eu}
}

\author{\IEEEauthorblockN{Claudio Gallicchio}
\IEEEauthorblockA{\textit{Department of Computer Science} \\
\textit{University of Pisa}\\
Pisa, Italy \\
gallicch@di.unipi.it}
}

\maketitle

\begin{abstract}
Reservoir Computing (RC) is a well-known strategy for designing Recurrent Neural Networks featured by striking efficiency of training. The crucial aspect of RC is to properly instantiate the hidden recurrent layer that serves as dynamical memory to the system. In this respect, the common recipe is to create a pool of randomly and \emph{sparsely} connected recurrent neurons. While the aspect of sparsity in the design of RC systems has been debated in the literature, it is nowadays understood mainly as a way to enhance the efficiency of computation, exploiting sparse matrix operations.\\
In this paper, we empirically investigate the role of sparsity in RC network design under the perspective of the richness of the developed temporal representations. We analyze both sparsity in the recurrent connections, and in the connections from the input to the reservoir. Our results point out that sparsity, in particular in input-reservoir connections, has a major role in developing internal temporal representations that have a longer short-term memory of past inputs and a higher dimension.
\end{abstract}


\section{Introduction}
Recurrent Neural Networks
(RNNs) \cite{kolen2001field} are a fundamental tool for adaptive processing of dynamically evolving information, with excellent performance in fields such as 
time-series forecasting \cite{laptev2017time},
machine translation \cite{sutskever2014sequence},
speech and text processing \cite{graves2013speech, nallapati2016abstractive}, just to mention a few.
An increasing number of works are analyzing the role of sparsity in the design of trained (dynamical) neural networks systems, for example through pruning \cite{narang2017exploring} or re-wiring \cite{bellec2017deep} connections. The characterization emerging from these studies is that having sparse connections between neurons
is not only advantageous in computational terms - as it enables fast sparse matrix computations - but can also be beneficial to obtain a better performance in practice.
Moreover, in the context of neurobiologically-inspired information processing systems, a sparse degree of connectivity between neurons has been shown to improve the quality of the developed internal representations \cite{litwin2017optimal}. Interestingly, the optimal amount sparsity in the numerical simulations matched observed properties of cerebellum-like circuits.

Reservoir Computing (RC) neural networks \cite{lukovsevivcius2009reservoir,schrauwen2007overview,jaeger2004harnessing} represent an intriguing development in the field of RNNs.
In RC, the recurrent hidden layer of a RNN is left untrained after initialization subject to asymptotic stability conditions of the corresponding dynamical system. As a result, learning is applied only to a simple readout component with striking advantages in terms of required training times compared to fully trained RNNs. Pushing the involved algorithms towards extreme simplicity and efficiency makes the RC approach very well suited for real-world application scenarios featured by (possibly severe) resource constraints, such as neuromorphic hardware implementations \cite{larger2012photonic} or cyber-physical systems where the learning modules are embedded at the edge \cite{bacciu2014experimental}.

A typical strategy in the design of RC networks is to setup the recurrent layer in a sparse way.
The initial intuition was that sparsity in the recurrent untrained layer could enable a decoupling of state variables and hence richer representations \cite{jaeger2001echo}.
Successively, several authors pointed out empirical evidences contrary to the initial intuition (see, e.g., 
\cite{schrauwen2007overview, xue2007decoupled, gallicchio2011architectural}). Currently, the sparse design of reservoirs is commonly understood mainly as a way to speedup state computations, without a practical effect on the resulting performance.
However, the impact of sparsity on the performance of RC neural networks has been typically studied limited to the recurrent connections only.
In this paper, we intend to shed more light on the role of sparsity in RC by extending the analysis to both recurrent and input connections.
Specifically, we empirically show the effect of recurrent and input sparsity in reservoirs, evaluated by means of short-term memory capacity and effective dimension of the resulting state trajectories.

The rest of this paper is structured as follows.
We introduce the basics of RC methodology in Section~\ref{sec.RC}, discussing initialization and sparsity of reservoirs. Then, in Section~\ref{sec.richness} we present the concepts of short-term memory capacity and effective reservoir dimension. Our experimental analysis is described in Section~\ref{sec.experiments}. Finally, in Section~\ref{sec.conclusions} we draw our conclusions and sketch possible developments.

\section{Reservoir Computing Neural Networks}
\label{sec.RC}
Here we give a brief description of the RC design methodology for RNNs, focusing on the Echo State Network (ESN) \cite{jaeger2004harnessing,jaeger2001echo} model.

\begin{figure}[htbp]
\centering
\includegraphics[width = 0.5\columnwidth]{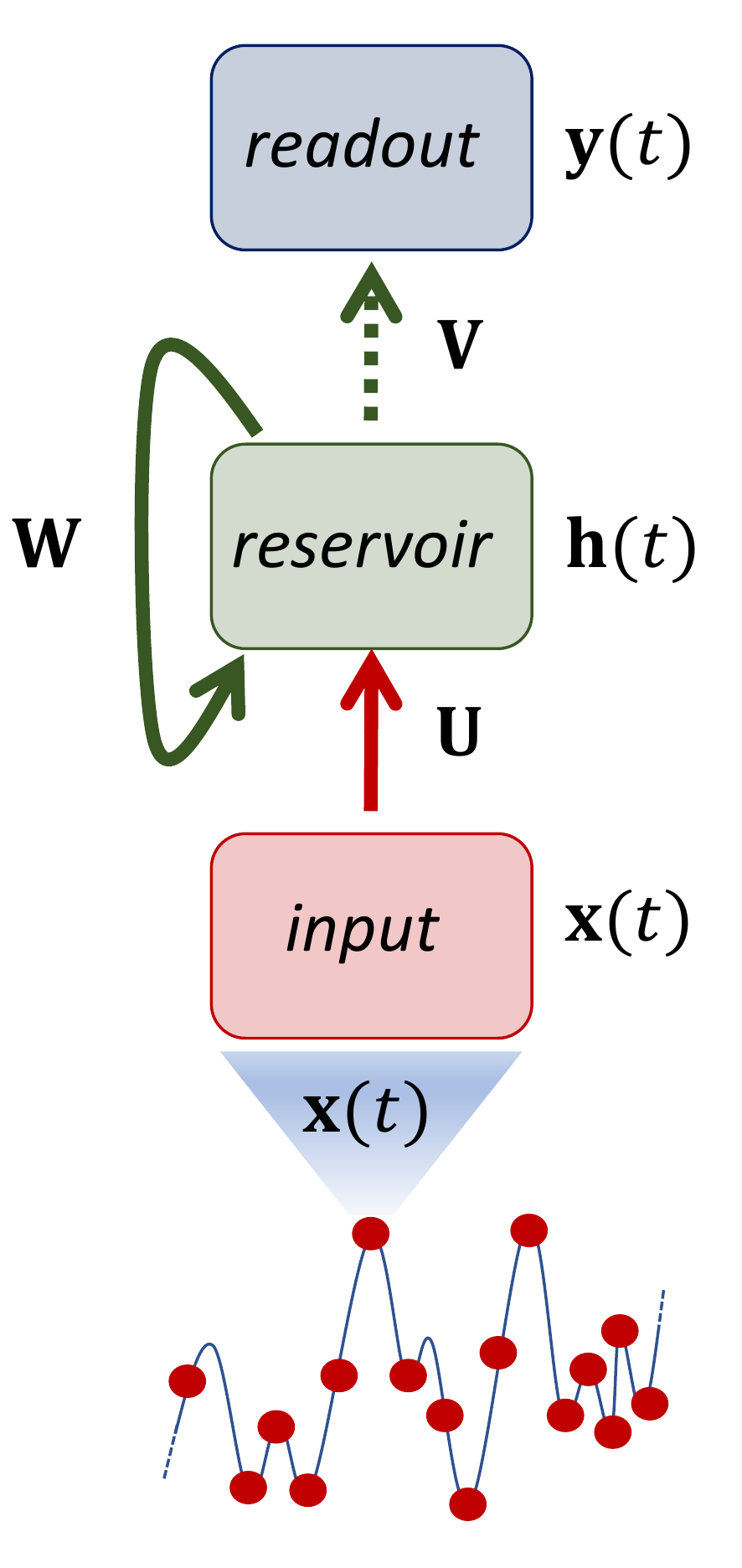}
\caption{Architecture of an RC neural network. The dotted arrow indicates trained connections.}
\label{fig.esn}
\end{figure}

An RC network is a neural information processing system that treats data in the form of (temporal) sequences. Architecturally, the neural network is composed by a hidden recurrent layer called \emph{reservoir}, and an output layer called \emph{readout}. Fig.~\ref{fig.esn} illustrates the building blocks of a typical RC network.

In what follows, we denote the number of reservoir neurons, i.e., the reservoir dimension, by $N$, and the state of the reservoir system at time $t$ by $\h(t) \in \R^{N}$.
This state is evolved by following a state update equation:
\begin{equation}
\label{eq.reservoir}
\h(t) = \tanh(\U \x(t) + \W \h(t-1)),
\end{equation}
where $\x(t) \in \R^{M}$ is the M-dimensional input at time $t$, $\U \in \R^{N \times M}$ is the input weight matrix, modulating the influence of the external input on the current state, and $\W \in \R^{N \times N}$ is the recurrent weight matrix, which controls the impact of previous state on the current state. The state is typically set to a zero vector as initial condition, i.e. $\h(0) = \mathbf{0} \in \R^{N}.$
Note that here we dropped from \eqref{eq.reservoir} the reference to bias terms to focus the analysis on the external stimulating input signal alone. Both weight matrices $\U$ and $\W$ remain untrained after initialization (see Section~\ref{sec.initialization}).

The reservoir system is coupled with a linear readout layer that computes an $L$-dimensional output at each time step, i.e. $\y(t) \in \R^{L}$, as an affine transformation of the reservoir state:
\begin{equation}
\label{eq.readout}
\y(t) = \V \h(t) + \mathbf{b},
\end{equation}
where $\V \in \R^{L \times N}$ is a readout weight matrix and $\mathbf{b} \in \R^{L}$ is a bias vector (that assumes a constant unitary input bias for the readout). The readout parameters are the only ones that undergo a training process, typically in closed-form fashion by using pseudo-inversion \cite{lukovsevivcius2009reservoir}.

\subsection{Initialization of Reservoirs}
\label{sec.initialization}
The fundamental characterization of RC neural networks is that all the reservoir parameters remain untrained after initialization. Such initialization is performed in agreement to asymptotic stability conditions expressed by the Echo State Property (ESP) \cite{yildiz2012re,jaeger2001echo,gallicchio2019chasing}, which essentially require to control the magnitude of the weights in $\U$ and $\W$. Usually, both the input weights in $\U$ and the recurrent weights in $\W$ are randomly drawn from a  uniform distribution in $[-1,1]$. After that, the elements in $\U$ are re-scaled by a factor $\omega_{in}$, which takes the role of input scaling. The weights in $\W$ are re-scaled to control the largest absolute eigenvalue, i.e., the spectral radius $\rho$, typically to a value smaller than 1 \cite{jaeger2001echo}.

The design strategy of the reservoir topology (i.e., the way in which the reservoir neurons are connected among each other) has been subject of several studies in literature (see, e.g., \cite{strauss2012design,rodan2010minimum}). While some of the proposed reservoir organizations can be beneficial in specific application circumstances, a random and sparse topological organization of the reservoir is the architecture of choice in general cases. This is the focus of our analysis in this paper.

\begin{figure}[htbp]
\centering
\includegraphics[width = 1\columnwidth]{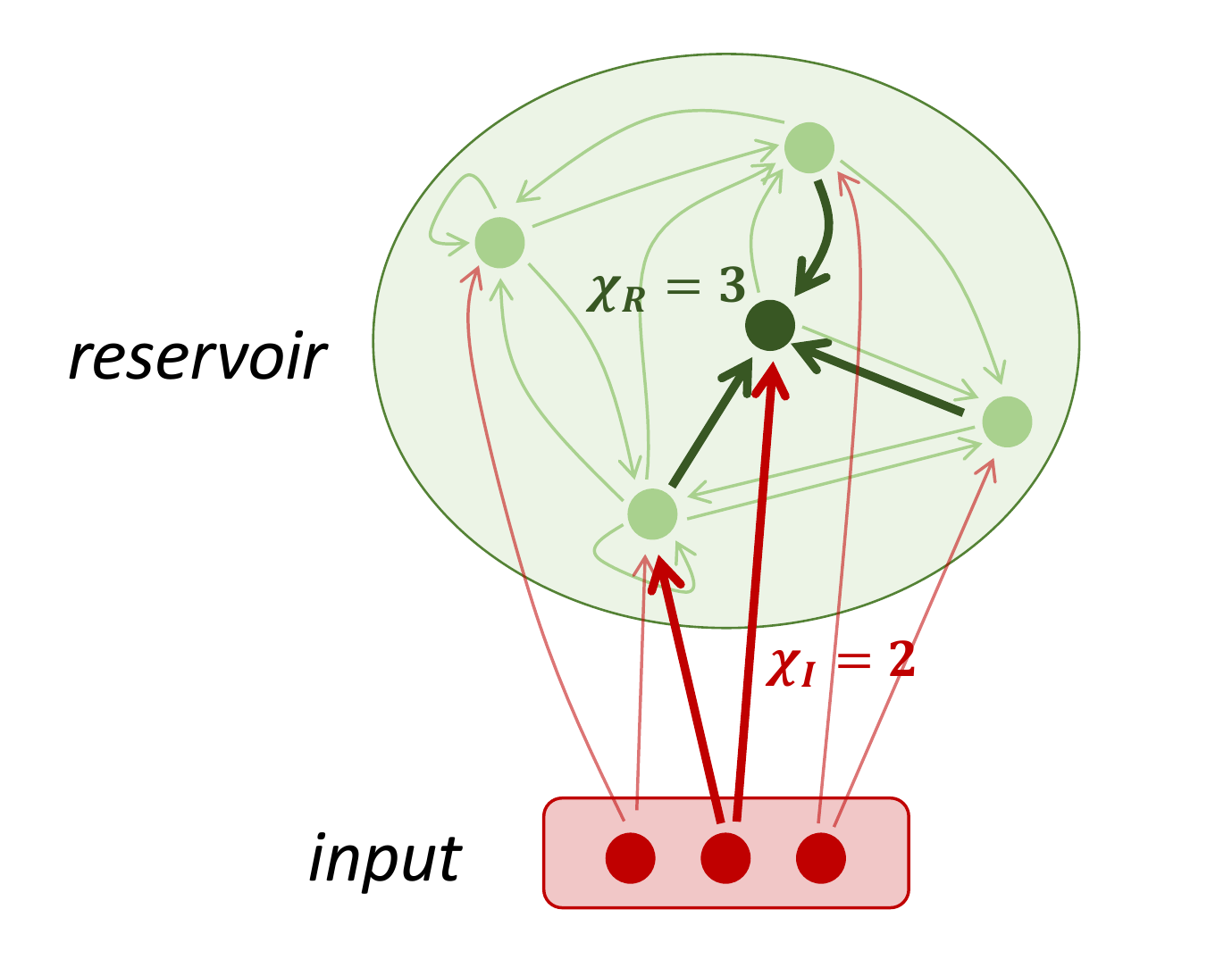}
\caption{Illustration of sparsity in input to reservoir and recurrent reservoir connections. $\chi_R$ indicates the number of incoming recurrent connections for each reservoir unit. $\chi_I$ indicates the number of outgoing connections from each input unit.}
\label{fig.sparsity}
\end{figure}

Making the connections among reservoir neurons sparse has the fundamental practical advantage to reduce the cost of state update operations in \eqref{eq.reservoir}. Actually, for densely connected reservoirs (and assuming $N >> M$) the cost of state updating scales as $\mathcal{O}(N^2)$, i.e. quadratically with the reservoir size. 
A first 
approach to make the reservoir sparsely connected would be to impose a (small) fixed percentage, say $C$, of non-zero weights in the involved weight matrices matrices. Although reducing the running times in practice for smaller reservoirs, this approach would asymptotically scale as $\mathcal{O}(N^2 \, C/100)$, hence still quadratically with the reservoir size. A more effective approach, which is adopted in this paper, is to fix the number, say $\chi_R$, of incoming recurrent connections for each reservoir unit. This indeed makes the state update cost as small as $\mathcal{O}(N \, \chi_R)$, i.e. scaling only linearly with the number of neurons in the reservoir.
A similar strategy can be adapted for the setup of the input connections. In this case, to ensure that each input dimension is actually forwarded to the reservoir, we fix the number of outgoing connections from each input units, denoted as $\chi_I$. The sparse architectural reservoir setup used in this paper is exemplified in Figure~\ref{fig.sparsity}. Notice that in this case, every row of $\W$ has exactly $\chi_R$ non-zero values, and every column of $\U$ has exactly $\chi_I$ non-zero elements, with both $\chi_R$ and $\chi_I$ being not greater than $N$.

\section{Short-term Memory and Effective Reservoir Space Dimension}
\label{sec.richness}
The role of the recurrent reservoir system is to embed the  input time-series into an internal ``state'' representation, given by the activation of the reservoir neurons over time. Here we analyze the quality of such internal reservoir representation by quantifying its short-term memory and effective dimension.\\

\noindent
\emph{Short-term Memory Capacity (MC)} \cite{jaeger2001short} tests the ability of a recurrent neural system to reconstruct its driving input time-series from the transient state dynamics. More in detail, the reservoir is driven by a uni-dimensional time-series, $x(t), t = 1, 2, \ldots$, and different readout units are trained to recall progressively delayed versions of the input. I.e., the i-th readout unit $y_i(t)$ should approximate $x(t-i)$. The MC of an RC network is then quantified as follows:
\begin{equation}
    \label{eq.mc}
    MC = \sum_{i = 1}^\infty \frac{cov^2(x(t-i),y_i(t))}{\sigma^2(x(t-i))
    \sigma^2(y_i(t))},
\end{equation}
i.e., as the sum of squared correlation coefficients of the delayed input and reconstructed signals.\\

\noindent
\emph{Effective Dimension ($N_{eff}$)}\cite{abbott2011interactions, litwin2017optimal} is a measure of the number of orthogonal directions in the neuronal system's state trajectory over time. While the evolution of the reservoir system in \eqref{eq.reservoir} is described by an $N$-dimensional state vector $\h(t)$, the actual reservoir trajectory lies into a lower-dimensional manifold whose dimension can be quantified as follows:
\begin{equation}
    \label{eq.dimension}
    N_{eff} = \frac{(\sum_{i = 1}^N \lambda_i)^2}{\sum_{i=1}^N\lambda_i^2},
\end{equation}
where $\lambda_i$, $i = 1, 2, \ldots, N$, denote the eigenvalues of the covariance matrix of the reservoir state activation over time. 
When measured for a reservoir under the driving influence of an external time-series, \eqref{eq.dimension} gives an estimate of the number of directions of reservoir state variability that are (linearly) uncorrelated along the observed trajectory.

\section{Experimental Analysis}
\label{sec.experiments}

We measured the short-term memory (MC) and the effective reservoir dimension ($N_{eff}$) introduced in Section~\ref{sec.richness} for RC networks varying the amount of recurrent and input connections.
Our experimental settings are described in Section~\ref{sec.settings}, while the results are reported in Section~\ref{sec.results}.

\subsection{Settings}
\label{sec.settings}
We used a uni-dimensional signal as driving input for the reservoir (i.e., $M = 1$). To maximally test the intrinsic  quality of reservoir representations, we used iid randomly sampled inputs $x(t)$ from a uniform distribution (in $[-0.8, 0.8]$).
The length of the generated input time-series was $6000$, and the number of reservoir neurons was fixed to $N = 100$.
To compute MC, we used the first $5000$ time-steps as training set\footnote{We used 
pseudo-inversion to train the readout, discarding the first 1000 time-steps as initial transient.}, using the remaining $1000$ time-steps to assess the MC score. The total number of delays used for the computation of \eqref{eq.mc} was 200, which is in practice sufficient to account for all the non-negligible contributions for 100-dimensional reservoirs. The last $1000$ time-steps of the dataset were also used to compute the effective reservoir dimension $N_{eff}$ (see \eqref{eq.dimension}).

In our experiments, we used RC networks with spectral radius $\rho = 0.9$ and input scaling $\omega_{in} = 1$. While this setup is of common use in RC practice, we also ran preliminary 
experiments with 
other choices of these hyper-parameters, finding that the outcomes are not qualitatively different. 
We varied both the number of recurrent connections ($\chi_R$) and of input connections ($\chi_I$) from 1 to 100 (with step of 1). For each configuration  we averaged the results over 50 reservoir realizations.

\subsection{Results}
\label{sec.results}

\begin{figure*}[htbp]
\centering
\begin{subfigure}{0.48\textwidth}
\centering
\includegraphics[width = \textwidth]{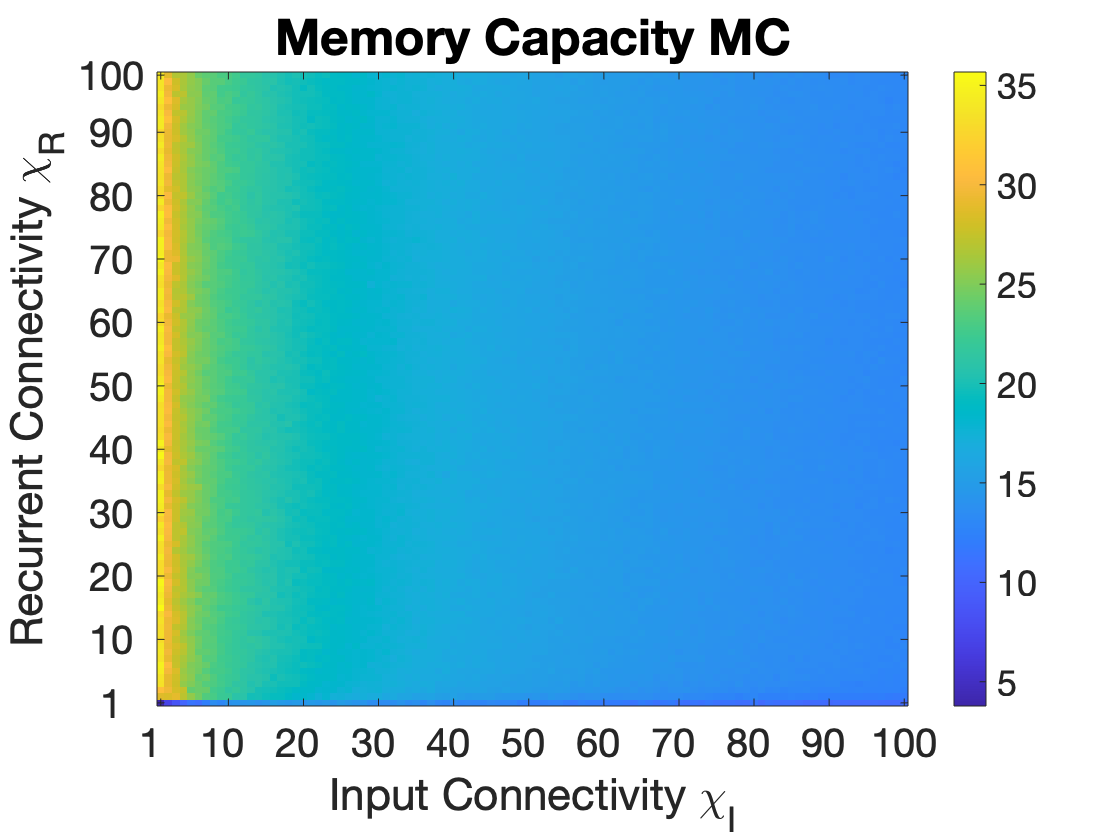}
\label{fig.MC}
\end{subfigure}
\begin{subfigure}{0.48\textwidth}
\centering
\includegraphics[width = \textwidth]{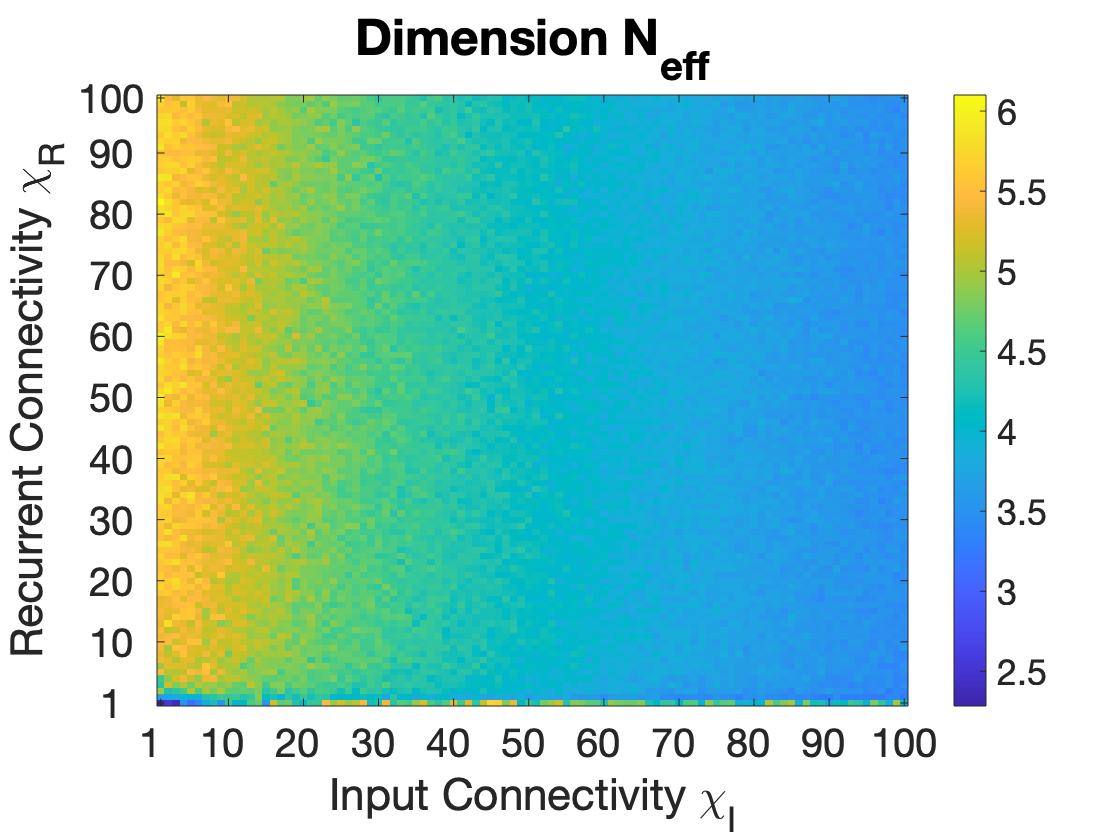}
\label{fig.dim}
\end{subfigure}
\caption{Short-term Memory Capacity (MC) and effective dimension ($N_{eff}$) of RC networks. Results corresponds to $N = 100$ reservoir neurons, spectral radius $\rho = 0.9$, and input scaling $\omega_{in} = 1$. Recurrent ($\chi_R$) and input ($\chi_I$) connectivity varied from 1 to 100 with step 1. For each of the 10000 configurations the results are averaged over a number of 50 reservoir realizations.}
\label{fig.results}
\end{figure*}

The achieved values of MC and $N_{eff}$ in correspondence of the possible sparsity settings (values of $\chi_R$ and $\chi_I$) are shown in Fig.~\ref{fig.results}.
We can draw two major observations from the results. First, the number of input connections has a decisive impact on both 
the short-term memory and the effective reservoir dimension of the networks.
Indeed, maximally sparse input connections, with $\chi_I = 1$, achieved the highest performances.
Interestingly, simply propagating the input to all the reservoir neurons degrades the performance sensibly. Second, the role of sparsity in recurrent connections seems to be much less important. In fact, the trend in Fig.~\ref{fig.results} indicates that for a given input connectivity, the achieved results are not much sensible to the exact number of recurrent connections (after a minimum number has been exceeded).

\begin{figure*}[htbp]
\centering
\begin{subfigure}{0.32\textwidth}
\centering
\includegraphics[width = \textwidth]{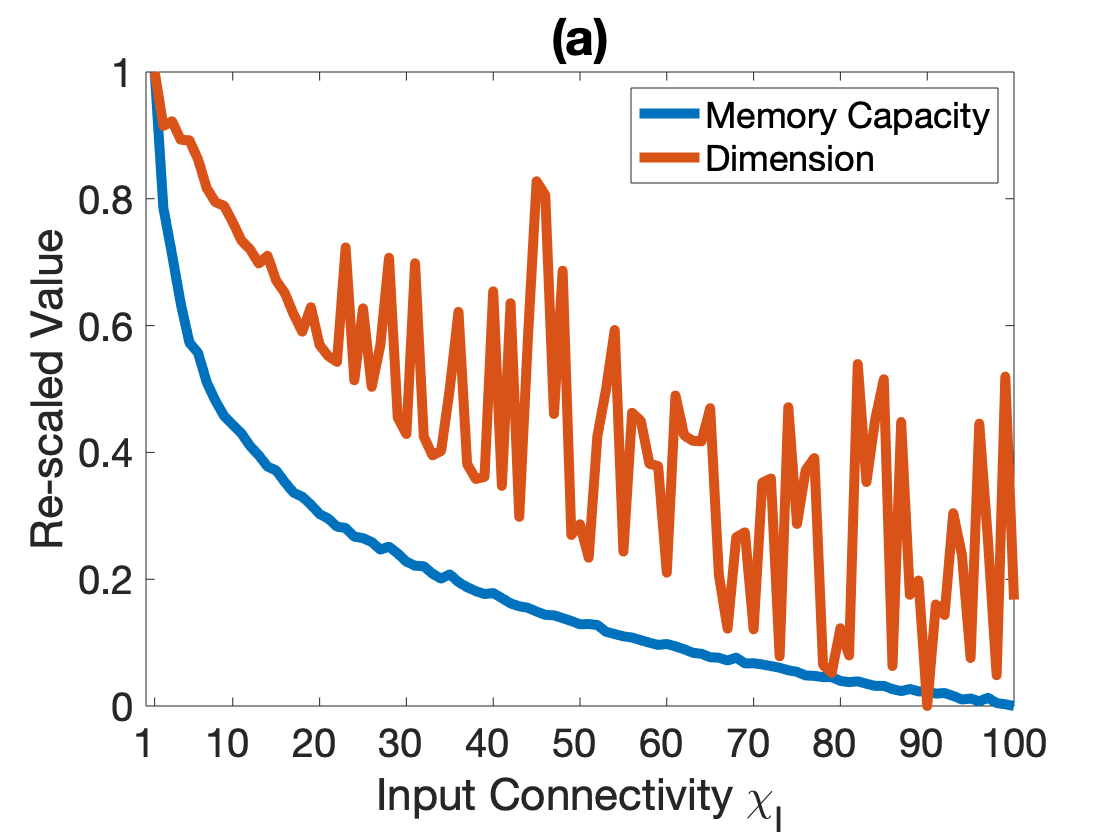}
\label{fig.dim}
\end{subfigure}
\begin{subfigure}{0.32\textwidth}
\centering
\includegraphics[width = \textwidth]{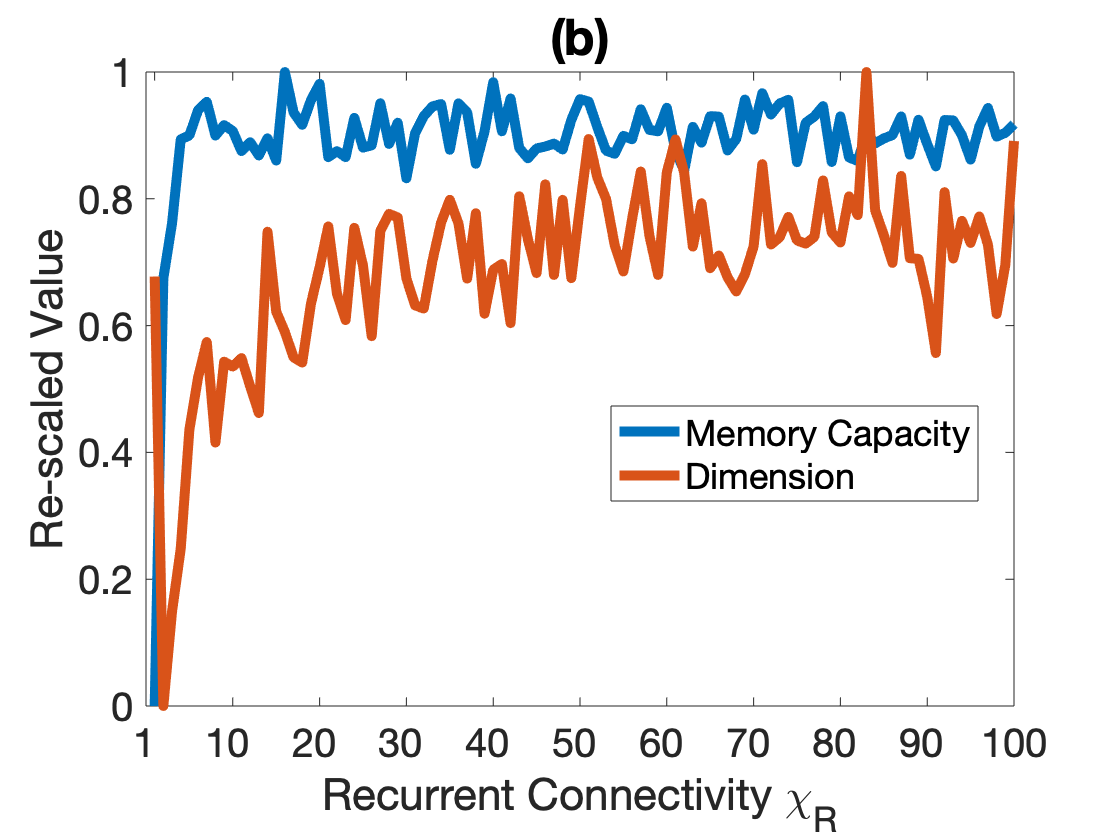}
\label{fig.dim}
\end{subfigure}
\begin{subfigure}{0.32\textwidth}
\centering
\includegraphics[width = \textwidth]{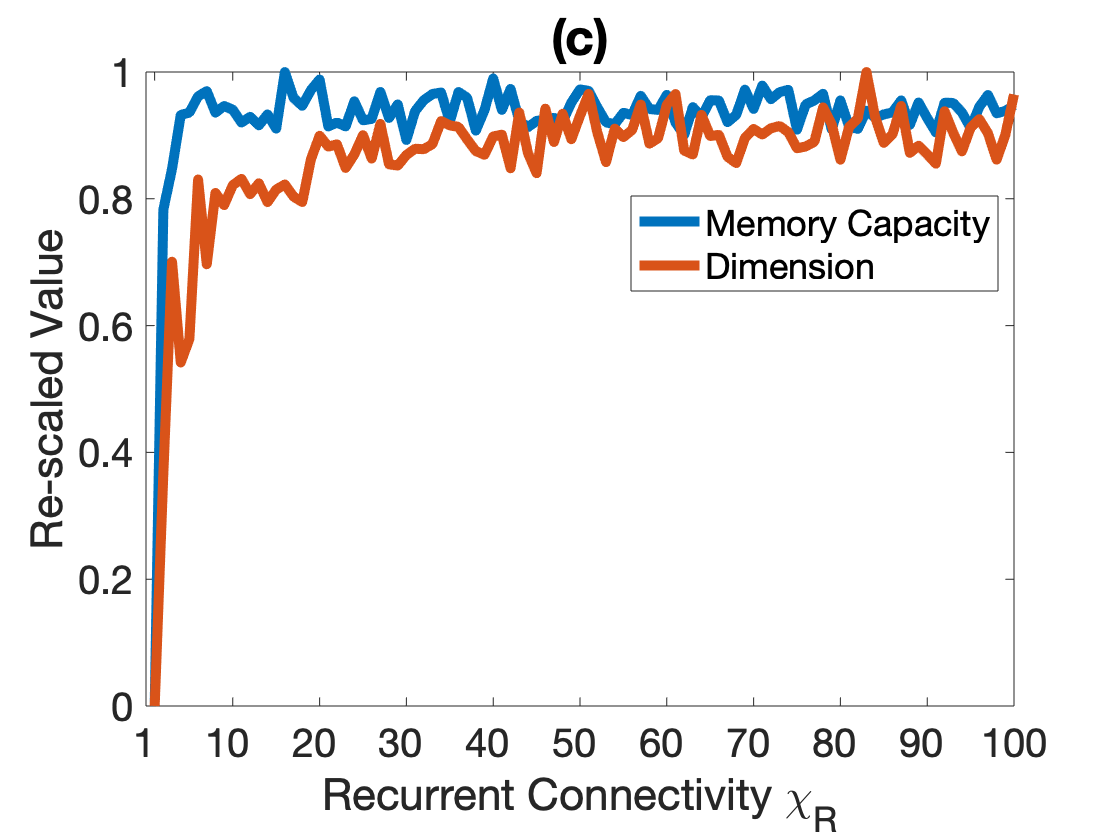}
\label{fig.MC}
\end{subfigure}
\caption{Short-term Memory Capacity (MC) and effective dimension ($N_{eff}$) of 100 units RC networks, detailed for:
\textbf{(a)} best results  for increasing input connectivity; 
\textbf{(b)} best results  for increasing recurrent connectivity; 
\textbf{(c)} results  for maximally sparse input connections ($\chi_I = 1$) and increasing recurrent connectivity.
Results are re-scaled to $[0,1]$.}
\label{fig.results2}
\end{figure*}

The results are further detailed in Fig.~\ref{fig.results2}, which shows
the best result for each choice of input (resp. recurrent) connectivity in Fig.~\ref{fig.results2}(a) (resp. Fig.~\ref{fig.results2}(b)), as well as
the results achieved for maximally sparse input connectivity, i.e. for $\chi_I = 1$, 
in Fig.~\ref{fig.results2}(c).
Figs.~\ref{fig.results2}(a)-(b) confirm the already observed trends. On the one hand the performance of the RC networks tends to deteriorate for less sparse input weight matrices. On the other hand, a modest number of recurrent connections is already sufficient to achieve a performance 
not far from
the highest possible one. For RC networks with $\chi_I = 1$ (Fig.~\ref{fig.results2}(c)), both MC and $N_{eff}$ saturate for fairly small values of $\chi_R$, without appreciable differences for settings with more than $20$ recurrent connections per reservoir neuron.

\section{Conclusions}
\label{sec.conclusions}
We have empirically analyzed the performance of RC neural networks in relation to sparsity of input and recurrent connections. Our results indicate that under commonly used reservoir configurations, the number of non-zero connections can play a decisive role in determining the richness of the developed representations. In particular, while a modest number of recurrent connections is already sufficient to achieve good performance, we found that maximally sparse input to reservoir connections lead to the best results both in terms of short-term memory and in terms of effective dimension of the 
state manifold. Overall, our analysis points out a simple rule of thumb for shaping reservoir weight matrices in case of uni-dimensional driving time-series: (i) connect the input to just one reservoir neuron, and (ii) 
set a small number of incoming recurrent connections ($\approx 20\%$) for each reservoir neuron.

The study presented in this paper can be seen as preparatory to opening further and deeper lines of research.
First of all, the role of sparsity 
can be investigated in synergy with \emph{structured} (rather than random) recurrent reservoir topologies, such as those based on cyclic \cite{rodan2010minimum} or small-world \cite{kawai2019small} connections.
Similarly, the study can be extended towards \emph{deep} RC neural networks \cite{gallicchio2017deep,gallicchio2018design}, where multiple reservoir layers are connected in a pipeline. In this case, the sparsity of input connections for higher layers has the even more intriguing role of modulating the extent of signal propagation between consecutive internal representations. Neuromorphic hardware implementations \cite{moughames2019three,freiberger2019towards,partzsch2011analyzing,larger2012photonic} of deep recurrent neural systems are an important example of a domain where such insights can be capitalized in practice.
Under a broader perspective, and outside the RC world, the analysis presented here pointed out that a sparse setting of RNN connections brings advantages even before learning of the non-zero connections. How these architectural advantages can be further exploited by (supervised or unsupervised) \emph{training} is another exciting open research question.

\bibliographystyle{IEEEtran}
\bibliography{references}

\end{document}